\pdfoutput=1
\documentclass[11pt,a4paper]{article}
\usepackage[hyperref]{emnlp2020}
\usepackage{times}
\usepackage{latexsym}

\usepackage{graphicx}
\graphicspath{ {.} }

\usepackage{microtype}

\aclfinalcopy 
\def\aclpaperid{2271} 

\title{Does the Objective Matter? 

Comparing Training Objectives for Pronoun Resolution}

\author{Yordan Yordanov\textsuperscript{1}, Oana-Maria Camburu\textsuperscript{1,2}, Vid Kocijan\textsuperscript{1}, Thomas Lukasiewicz\textsuperscript{1,2}\\
  \textsuperscript{1}University of Oxford, Oxford, UK \\
  \textsuperscript{2}Alan Turing Institute, London, UK \\
  \texttt{firstname.lastname@cs.ox.ac.uk}}

\date{}

\usepackage{amsmath}

\begin{document}
\maketitle
\begin{abstract}

Hard cases of pronoun resolution have been used as a long-standing benchmark for commonsense reasoning. In the recent literature, pre-trained language models have been used to obtain state-of-the-art results on pronoun resolution. Overall, four categories of training and evaluation objectives have been introduced. The variety of training datasets and pre-trained language models used in these works makes it unclear whether the choice of training objective is critical. In this work, we make a fair comparison of the performance and seed-wise stability of four models that represent the four categories of objectives. 
Our experiments show that the objective of sequence ranking performs the best in-domain, while the objective of semantic similarity between candidates and pronoun performs the best out-of-domain. We also observe a seed-wise instability of the model using sequence ranking, which is not the case when the other objectives are used.
\end{abstract}

\section{Introduction}

Hard cases of pronoun resolution have been a long-standing problem in natural language processing, which has served as a performance benchmark for the research community \cite{levesque2012winograd, wang2018glue, wang2019superglue}. 
For example, the WinoGrande dataset \cite{WinoGrande} consists of pronoun resolution schemas that are constructed so that resolving them requires background knowledge and commonsense reasoning. In WinoGrande, the pronoun is obscured by ``\underline{\hspace{0.5cm}}'' to remove gender and number cues. The task is to find the correct candidate for ``\underline{\hspace{0.5cm}}'' out of two given candidates. For example:
\begin{quote} 
\textit{John moved the couch from the garage to the backyard to create space. The \underline{\hspace{0.5cm}} is small.}
Candidates: garage, backyard.
\end{quote}

Recently, supervised learning on top of pre-trained language models has been established as the main approach for pronoun resolution \cite{Trick, WikiCREM, WinoGrande}. Under this type of approach, we identify four categories of objectives commonly used for pronoun resolution: 
\begin{enumerate}
    \item \label{itm:first} comparing the language model probabilities for each candidate \cite{Trick, WikiCREM, HNN},
    \item \label{itm:second} using semantic similarity between the pronoun and the candidates \cite{UDSSM, HNN},
    \item \label{itm:third} using sequence ranking among the possible substituted sentences \cite{sequence-ranking, WinoGrande}, and 
    \item \label{itm:fourth} selecting a candidate based on the attentions of the pronoun in a transformer model \cite{attention-not-all}.
\end{enumerate}

We list one representative model from each category.
For \ref{itm:first}, \citet{Trick} use the BERT masked language model \cite{devlin2018bert} to produce the probabilities of the pronoun to be replaced with each of the two candidates. For \ref{itm:second}, the Unsupervised Deep Structured Semantic Model (UDSSM-I) \cite{UDSSM} uses contextualized word embeddings produced by a bidirectional recurrent neural network (BiRNN), and then compares the word embedding of each candidate with the word embedding of the pronoun. For \ref{itm:third}, RoBERTa-WinoGrande \cite{WinoGrande} encodes a pair of sentences (one for each candidate substituted in the input) by using RoBERTa \cite{Roberta} to determine which substitution is the correct one. Finally, the zero-shot Maximum Attention Score (MAS) model \cite{attention-not-all} selects a candidate based on how much the pronoun attends to each candidate internally in BERT. 

The problem with all these objectives is that they have not been introduced under the same circumstances. They use different language models and word embeddings (e.g., BERT, RoBERTa, or BiRNN), and have been trained on different data (e.g., DPR \cite{DPR}, WinoGrande, or no additional data). Therefore, it is unclear whether the choice of the objective function is essential for pronoun resolution tasks.
Moreover, the seed-wise stability and the expected performance of these models have usually not been reported. However, seed-wise instability and performance variation are well-known problems when fine-tuning transformer-based models \cite{liu2020understanding, dodge2020finetuning}.

In this work, we compare the performance and seed-wise stability of the four categories of training objectives for pronoun resolution on equal grounds. To do this, for category \ref{itm:fourth}, we adapt to training the zero-shot MAS model. For category~\ref{itm:second}, we also introduce Coreference Semantic Similarity (CSS), which is a simplification and modification of UDSSM-I for transformer encoders.
We select WinoGrande as our training and development dataset due to its large size (40,938 examples) and generalizability to other pronoun resolution tasks \cite{WinoGrande}. We also use for testing the following well-established datasets: the Winograd Schema Challenge dataset (WSC) \cite{levesque2012winograd} and the Definite Pronoun Resolution dataset (DPR) \cite{DPR}.
We choose as language model RoBERTa \cite{Roberta}, as it significantly outperforms BERT on WinoGrande, WSC, and DPR \cite{WinoGrande}.

Finally, our evaluations are done under an unprecedentedly large number of seeds (20).

\section{Models}

This section presents the four training objectives and the models\footnote{The code is publicly available at: \url{https://github.com/YDYordanov/WS-training-objectives}.} that represent each of them.

All four models share the RoBERTa\footnote{\textit{roberta-large} from \cite{Wolf2019HuggingFacesTS}} contextualized word embeddings. RoBERTa has an identical transformer architecture to BERT \cite{devlin2018bert}, with the only difference being the training procedure. Hence, RoBERTa is a masked language model that outputs the probability distribution for filling a gap in the text (denoted by a ``\textless mask\textgreater \hspace{0.5pt}'' token).
Additionally, RoBERTa is a text encoder, with one output for each token of the input sentence. Three of the models (\ref{WG-baseline}, \ref{CSS}, and \ref{MAS}) use a multi-layer perceptron (MLP) classification ``head'', which takes some part of the encoder as input. 

All four models use binary cross-entropy loss with a pair of probabilities as input, and the following target labels: sentence correctness for \ref{WG-baseline} and candidate correctness for \ref{BWP}, \ref{CSS}, and \ref{MAS}. 

\subsection{WinoGrande Sequence Ranking} \label{WG-baseline}

We refer to the RoBERTa-WinoGrande model introduced by \citet{WinoGrande} as WG-SR, since it has a sequence ranking objective.
This model predicts which sentence of a pair of substituted sentences is more plausible. Each of the pair of sentences in the input of WG-SR is split in two before the substituted candidate. For example,
\begin{quote} 
\textless s\textgreater \hspace{0.5pt} The city councilmen refused the demonstrators a permit because \textless /s\textgreater \hspace{0.5pt} \textless /s\textgreater \hspace{0.5pt} \underline{\hspace{0.5cm}} feared violence. \textless /s\textgreater, 
\end{quote} 
where ``\underline{\hspace{0.5cm}}'' is filled with each of the two candidates: ``the city councilmen'' or ``the demonstrators''.

The WG-SR code\footnote{\url{https://github.com/allenai/winogrande}} is based on the RobertaForMultipleChoice model \cite{Wolf2019HuggingFacesTS}, restricted to binary choice. This model consists of the pre-trained RoBERTa encoder and an MLP head based on the \textless s\textgreater \hspace{0.5pt} (first) token of RoBERTa's output. The MLP has one hidden layer with tanh activation, hidden size matching that of the encoder, and one-dimensional output. The pair of input sentences $(\text{S}_1, \text{S}_2)$ thus produces a pair of values, which are then passed through a softmax to obtain the two sentence probabilities $P(\text{S}_1)$ and $P(\text{S}_2)$.

\subsection{Binary Word Prediction} \label{BWP}

We denote by Binary Word Prediction (BWP) the model suggested by \citet{Roberta} in their code repository\footnote{\url{https://github.com/pytorch/fairseq/tree/master/examples/roberta/wsc}} as a modification of the model from \citet{Trick}. Instead of using margin loss, BWP uses binary cross-entropy loss. We select this modified version, because it is claimed to be more robust by its authors, and it also has two fewer hyperparameters. 

For a given (unsubstituted) input sentence, the BWP model estimates which of the two candidates is more likely to fill the gap ``\underline{\hspace{0.5cm}}''. The input format is like in the following example, where ``\underline{\hspace{0.5cm}}'' is replaced by the ``\textless mask\textgreater''
token, to serve for the masked language model: 
\begin{quote}
\textless s\textgreater \hspace{0.5pt} The city councilmen refused the demonstrators a permit because \textless mask\textgreater \hspace{0.5pt} feared violence. \textless /s\textgreater
 \end{quote}
 
With such an input, the RoBERTa masked language model returns the log-probability predictions at the ``\textless mask\textgreater \hspace{0.5pt}'' token over the vocabulary. Of those predictions, only the ones corresponding to the two word candidates $c_1$ and $c_2$ are selected by BWP: $\text{log}P_{\text{vocab}}(c_1)$ and $\text{log}P_{\text{vocab}}(c_2)$. Here, the log-probability of each candidate is defined by averaging the log-probabilities of its tokens. Then, softmax is computed with inputs $\text{log}P_{\text{vocab}}(c_1)$ and $\text{log}P_{\text{vocab}}(c_2)$, which is how we define the pair of probabilities: $(P(c_1), P(c_2)) := (P_{\text{vocab}}(c_1) / (P_{\text{vocab}}(c_1) + P_{\text{vocab}}(c_2)), P_{\text{vocab}}(c_2) / (P_{\text{vocab}}(c_1) + P_{\text{vocab}}(c_2)))$.

\subsection{Coreference Semantic Similarity} \label{CSS}

We propose Coreference Semantic Similarity (CSS), a modification of the training objective of the Unsupervised Deep Structured Semantic Model (UDSSM-I) \cite{UDSSM}. Like UDSSM-I, the CSS objective works by comparison in the word embedding space, such that the candidate that is more similar to the embedding of the pronoun is selected. Unlike UDSSM-I, the CSS objective is simpler, with no attention weights on the tokens of the candidates. It also uses a transformer encoder instead of a recurrent neural network, which enables it to take advantage of state-of-the-art pre-trained language models. 

The input format for this model is the same as for BWP (\ref{BWP}). This input is used by RoBERTa to produce contextualized word embeddings. For each candidate $c$, we define its contextualized word embedding $\text{emb}(c)$ by averaging the contextualized word embeddings of its tokens. 

For classification, we compare the similarity scores of the embeddings of the \textless mask\textgreater \hspace{0.5pt} token with each of the two candidates $c_1$ and $c_2$, i.e., we compare $\text{sim}(\text{emb}(c_1), \text{emb}(\text{\textless mask\textgreater)})$ and $\text{sim}(\text{emb}(c_2),$  $\text{emb}(\text{\textless mask\textgreater)})$ and select the candidate with greater similarity. 

For the similarity score function, we use \textit{additive alignment} \cite{bahdanau2014neural}, i.e., $\text{sim}(x, y) : = v^\top \text{tanh}(Wx+Uy)$, with the trainable parameters: vector $v$, and matrices $W$ and $U$, with hidden size equal to that of RoBERTa and output size of one.

During training, $\text{sim}(\text{emb}(c_1), \text{emb}(\text{\textless mask\textgreater}))$ and $\text{sim}(\text{emb}(c_2), \text{emb}(\text{\textless mask\textgreater}))$ are fed to a binary softmax function to obtain $P(c_1)$ and $P(c_2)$.

\subsection{Maximum Attention Score} \label{MAS}

The Maximum Attention Score (MAS) model was originally developed for zero-shot evaluation of transformer models on pronoun disambiguation \cite{attention-not-all}. It uses the attentions of all layers of a transformer model to produce a maximum attention score for each candidate that summarizes how much the pronoun attends to a candidate. The candidate that is most attended is selected. We adapt this objective to be trainable by replacing the summary of attentions with an MLP over the concatenated masked attention tensors, followed by a binary classifier.

The input of MAS is the same as for BWP (\ref{BWP}). Then, similarly to \citet{attention-not-all}, we extract the two attention tensors $A_{c_1}$ and $A_{c_2}$ given by the multi-layer RoBERTa attentions of the ``\textless mask\textgreater'' token to each of the two candidates $c_1$ and $c_2$, respectively. For each candidate $c$, the attention tensor $A_{c}$ is defined as the average of the attention tensors of all tokens that form $c$. The two corresponding max-masking tensors $M_{c_1}$ and $M_{c_2}$ are then derived as follows: for $i =1, 2$ and for each multi-index $j$ of the tensor $A_{c_i}$, we set $M_{c_i}(j) = 1$, if $A_{c_i}(j) \ge A_{c_{3-i}}(j)$, and $M_{c_i}(j) = 0$, otherwise. We obtain the two corresponding max-masked tensors by the element-wise products: $B_{c_1} = A_{c_1} \circ M_{c_1}$ and $B_{c_2} = A_{c_2}\circ M_{c_2}$.

Unlike \citet{attention-not-all}, we introduce an MLP on top of the concatenated tensor $B = [B_{c_1}, B_{c_2}]$ for binary classification. The MLP has two hidden layers, tanh activation, hidden size the same as its input, and two-dimensional output. It is followed by a binary softmax function to produce the two candidate probabilities $P(c_1)$ and $P(c_2)$.

\begin{table*}
\centering
\begin{tabular}{llll}
\hline
\textbf{Model} & \textbf{WG-dev} & \textbf{WSC} & \textbf{DPR}\\
\hline
WG-SR & \textbf{78.2} (1.00) & 89.2 (1.12) & 92.2 (0.61) \\
BWP & 76.3 (0.5) & 89.6 (0.80) & 91.8 (0.55) \\
CSS & 77.4 (0.78) & \textbf{90.2} (0.90) & \textbf{92.7} (0.66) \\
MAS & 76.6 (0.77) & 89.0 (1.51) & 92.3 (0.71) \\
\hline
\end{tabular}
\caption{\label{seeds_table}
Seed-wise aggregated performance of models on WG-dev, WSC, and DPR. The number format is: average accuracy in \%, and standard deviation (in parentheses). Out of the 20 seeds, only the converging ones are included. The best performance is marked in bold.
}
\end{table*}

\begin{table*}
\vspace*{0.7 cm}
\centering
\begin{tabular}{lcccc}
\hline
\textbf{Model} & \textbf{Maximum} & \textbf{Average} & \textbf{Standard deviation} & \textbf{Number of converged} \\
\hline
WG-SR & \textbf{80.0} & \textbf{76.8} & 2.28 & 49 out of 96 \\
BWP & 77.6 & 75.4 & 1.45 & 54 out of 96 \\
CSS & 78.9 & 76.2 & 1.13 & 56 out of 96 \\
MAS & 77.7 & 74.5 & 2.50 & 69 out of 96 \\
\hline
\end{tabular}
\caption{\label{hyperparam_wise_results_table}
Performance of all four models on WG-dev aggregated across all 96 hyperparameter combinations (including the three seeds). The numbers in the first three columns are: maximum accuracy in \%, average accuracy in \%, standard deviation. Only the converging models (with at least 60\% accuracy) are reported, and their number is in the last column.  The best performance is marked in bold.
}
\end{table*}

\section{Experiments}

For all four models, we select the best hyperparameters via grid search using 3 seeds, and then train the models with the best hyperparameters on 20 additional seeds. For WinoGrande, we use WG-dev (1,267 examples) for selecting the hyperparameters, and WG-train-XL as our training dataset. Due to the submission limitation (maximum one per week) of the WinoGrande leaderboard,\footnote{\url{https://leaderboard.allenai.org/winogrande/submissions/public}} we are unable to report all 80 trained models on WG-test, and instead we report them on WG-dev. For additional verification, we include results over the hyperparameter space, where WG-dev is a true test set.
We also report all models on the out-of-domain pronoun resolution datasets WSC (273 examples) and DPR (564 examples). The candidates provided in WSC were treated differently for the CSS and MAS models, as these models require precise candidate localization (see Appendix \ref{sec:wsc_preprocessing}).

For all four models, we do a grid search over the learning rate $\{5e-6, 1e-5, 3e-5, 5e-5\}$, the number of training epochs $\{3, 4, 5, 8\}$, and the batch-size $\{8, 16\}$, and we run each model with three different random seeds. This hyperparameter space is selected based on the union of the grid search by the original WG-SR work \cite{WinoGrande} and our observations on the other three models. The best hyperparameters (in Appendix \ref{sec:hyperparams}) are selected based on the maximum WG-dev accuracy across the three seeds. 

For all experiments, we use linear learning rate decay with warm-up over 10\% of the training data, and the AdamW optimizer \cite{Wolf2019HuggingFacesTS}, for which we only alter the learning rate.

\section{Results}

Table~\ref{seeds_table} shows the final seed-wise results for all four objectives. We see that the semantic similarity objective (CSS) outperforms the other three objectives on out-of-domain testing, with 90.2\% average accuracy on WSC and 92.7\% average accuracy on DPR. On the other hand, the sentence ranking objective used by WG-SR clearly outperforms the other three objectives on in-domain testing, with 78.2\% average accuracy on WG-dev. This is confirmed by the contents of Table~\ref{hyperparam_wise_results_table}, where we see that WG-SR has a better mean and max accuracy on WG-dev over the entire hyperparameter space compared to the other three models. For these cases, WG-dev is a true test set,  since early stopping was not used, and all tested setups are reported; hence, WG-dev has not influenced the models reported in Table~\ref{hyperparam_wise_results_table}.

In order to verify the statistical significance of our main results, we used the t-test for similar variances and different sample sizes to compare the distributions of accuracy on the converging seeds. Comparing the accuracies of CSS and WG-SR on WG-dev, WSC, and DPR, respectively, we get the following two-tailed $p$-values: $0.008249$, $0.003026$, and $0.017441$. All results are significant with $ p < 0.05 $.

We also observe that, even with the best hyperparameter combination, WG-SR exhibits seed-wise instability, as it fails to converge on 2 out of 20 seeds. This does not happen to the other three models. After considering 10 additional seeds, we obtained that WG-SR fails to converge on 10\% of the seeds (3 out of 30). 

Moreover, during the hyperparameter search, we observed that all models were prone to not converge for certain combinations of hyperparameters. The convergence threshold that we used was selected as having $\le 60\%$  accuracy on WG-dev, and its value was selected based on the performance distribution of all models. We observed that all models either perform around $50\%$ accuracy or $70\%$ accuracy or more on WG-dev. $60\%$ in this context is a good middle ground threshold. Table~\ref{hyperparam_wise_results_table} shows that
MAS converged most often; however, it also had the highest performance variation with a standard deviation of 2.5. Out of the four models, WG-SR converged least often, for only 49 out of all 96 hyperparameter combinations.

WG-SR likely performs better in-domain than CSS, MAS, and BWP, since those three use existing properties of RoBERTa (such as the possibility to compare contextualized embeddings, the attention structure of the model, and its pre-trained LM prediction head, respectively) for a task that they were not originally designed for (pronoun resolution). WG-SR, on the other hand, only uses the output of RoBERTa at the 0-th token, which is not pre-trained.

We identify two possible reasons why WG-SR performs worse than CSS on out-of-domain examples. The first reason is the one mentioned above, namely, not explicitly exploiting the listed properties of the pre-trained model would lead to a better fit on a specific dataset, but worse ``general knowledge''. This reason is not completely warranted, since WG-SR has similar out-of-domain performance to BWP and MAS. 
The second possible reason is that CSS uses an explicit candidate localization and candidate-pronoun matching (by comparing the embedding of the candidate and the pronoun), whereas in WG-SR these are achieved implicitly by feeding a pair of sentences to the model, one with the correct and one with the incorrect substitution. Again, this reason is not completely warranted, since MAS also uses explicit candidate localization and candidate-pronoun matching, but has a similar out-of-domain performance to WG-SR. Further investigation on the reasons why CSS outperforms WG-SR on the out-of-domain examples is left for future work.

\section{Summary and Outlook}

In this work, we categorized four existing objectives for pronoun resolution, and compared their performance and seed-wise stability on equal grounds. Our experiments showed that, on in-domain testing, the objective of sequence ranking based on the first token in RoBERTa outperforms the other three objectives, but can exhibit convergence problems. On out-of-domain testing, the objective of semantic similarity between the pronoun and each candidate outperforms the other three objectives.

Future work may investigate whether these results translate to other language models besides RoBERTa as well as other training datasets besides WinoGrande. Also, one could analyze the strengths and weaknesses of each objective, and evaluate other variations of these objectives.

\section*{Acknowledgments}
This work was supported by a JP Morgan PhD Fellowship, the Alan Turing Institute under the EPSRC grant EP/N510129/1, the AXA Research Fund, the ESRC grant ``Unlocking the Potential of AI for Law'', the EPSRC studentship OUCS/EPSRC-NPIF/VK/1123106, and EU Horizon 2020 under the grant 952215.
We also acknowledge the use of the EPSRC-funded Tier 2 facility JADE (EP/P020275/1).

\bibliography{anthology,emnlp2020}
\bibliographystyle{acl_natbib}

\clearpage
\appendix

\section{Best Hyperparameters}\label{a:best_hyp}
\label{sec:hyperparams}

See Table~\ref{best_hyperparams} for the best hyperparameters for each model.

\begin{table}
\centering
\begin{tabular}{lccc}
\hline
\textbf{Model} & \textbf{epochs} & \textbf{batch size} & \textbf{learn. rate}\\
\hline
WG-SR & 5 & 16 & 1e-5 \\
BWP & 8 & 16 & 1e-5 \\
CSS & 8 & 16 & 1e-5 \\
MAS & 8 & 8 & 1e-5 \\
\hline
\end{tabular}
\caption{\label{best_hyperparams}
The best hyperparameters for every model. 
}
\end{table}

\section{WSC Preprocessing}\label{a:wsc}
\label{sec:wsc_preprocessing}

When evaluating the CSS and the MAS model on the WSC dataset, we noticed a problem with the dataset, which interfered with locating the candidates in the text. The problem is that, in some WSC examples, the given candidate options do not match word-by-word the candidates as they appear in the text. For example,

\begin{quote}{\textit{Madonna fired her trainer because \underline{\hspace{0.5cm}} couldn't stand her boyfriend.}

Candidates: Madonna, The trainer.}
\end{quote}

In this example, we resolve this problem by manually replacing the candidate option ``the trainer" with ``her trainer", to match exactly the candidate as it appears in the text. By following this procedure, we manually modified all 88 problematic examples in WSC (out of 273 examples in total). Note that this problem does not exist for WinoGrande and DPR. Furthermore, in real-world applications, such a problem does not exist, since the candidates are not provided and have to be extracted automatically from the text. Detected candidates thus match the spans in the text.

We use this modified version of WSC only for the CSS and MAS models, because they require precise candidate localization. For WG-SR and BWP, we use the unmodified WSC version. The edited dataset can be found in the code repository\footnote{\url{https://github.com/YDYordanov/WS-training-objectives}}.

\end{document}